\documentclass{article}

\usepackage[preprint]{neurips_2026}
\usepackage{xspace}    
\usepackage{subcaption}
\usepackage{booktabs,makecell,threeparttable,multirow}

\usepackage[utf8]{inputenc} %
\usepackage[T1]{fontenc}    %
\usepackage{hyperref}       %
\usepackage{url}            %
\usepackage{booktabs}       %
\usepackage{amsmath}        %
\usepackage{amsfonts}       %
\usepackage{nicefrac}       %
\usepackage{microtype}      %
\usepackage{xcolor}         %
\usepackage{graphicx}       %
\graphicspath{{./}{./figures/}{./neurips/}{./neurips/figures/}}
\usepackage{cleveref}
\usepackage{enumitem}

\newcommand{\name}{DiscoLoop\xspace}

\title{DiscoLoop: Looping Discrete Embeddings and Continuous Hidden States for Multi-hop Reasoning }

\author{Hengyu Fu$^1$\thanks{Equal Contribution.} \And Tianyu Guo$^1$\footnotemark[1] \And Zixuan Wang$^{1,2}$\footnotemark[1]\And Hanlin Zhu$^1$\footnotemark[1] \AND Jason D. Lee$^1$\And Jiantao Jiao$^1$\And Stuart Russell$^1$\And Song Mei$^1$}

\begin{document}

\maketitle
\footnotetext[1]{University of California, Berkeley. Email: \texttt{\{hengyuf,tianyu\_guo,hanlinzhu\}@berkeley.edu}}
\footnotetext[2]{Princeton University. Email: \texttt{zw2814@princeton.edu}}

\begin{abstract}
Large language models achieve strong performance on many reasoning tasks when allowed to externalize intermediate steps as Chain-of-Thought (CoT). However, many questions require the model to internalize the \textit{multi-step reasoning} within a single forward pass before generating the answer. We study this challenge through \textit{two-hop reasoning}, a representative task where the model must compose multiple pieces of parametric knowledge within a single forward pass. Standard non-recurrent Transformers suffer from a \textit{depth-local storage} problem: facts learned in earlier layers are unavailable where second-hop retrieval happens. We found that Looped Transformers mitigate this issue by reusing the same memory, but still generalize imperfectly.
We show that the remaining bottleneck is \textit{representational}. In the two-hop reasoning task, the first loop often makes the correct bridge entity nearly perfectly decodable, yet the corresponding hidden state remains poorly aligned with the bridge token embedding. Surprisingly, an easy training-free realignment intervention nearly closes the generalization gap.
 Building upon this insight, we propose \textbf{\name}, a looping architecture whose recurrence carries both a \emph{\textbf{dis}}crete embedding channel and a \emph{\textbf{co}}ntinuous hidden-state channel. \name~achieves near-perfect accuracy with substantially fewer training steps across symbolic and synthetic-language multi-hop reasoning tasks. When applied to real-world pretraining, \name~attains lower training loss and stronger benchmark performance than looped-transformer baselines, suggesting that the mixed-channel design transfers to practical language modeling.

\end{abstract}

\section{Introduction}
\label{sec:intro}

Current large language models (LLMs) heavily rely on long traces of Chain-of-Thought (CoT)~\citep{wei2022chain} to achieve strong performance in various challenging tasks, including math~\citep{jaech2024openai,guo2025deepseek} and coding~\citep{chen2021evaluating,cao2026qwen3}. Yet many questions implicitly require several reasoning steps before generating the answer token, which must happen \textit{internally} through the forward pass~\citep{yang2024large}. Ideally, an intelligent model should reason in a parameter and context-efficient way by \textit{storing} atomic facts and \emph{composing} them implicitly on demand, which reduces the need to verbalize every intermediate step in the context. Unfortunately, many recent works show that transformer-based LLMs struggle with such \textit{implicit reasoning} \citep{ye2024physics,press2023measuring, yang2025large} despite their excellent performance with CoT.

Towards understanding the failure on implicit reasoning, a symbolic \textit{two-hop reasoning} task \citep{wang2024grokked,ye2025does} was proposed as the most fundamental version of implicit reasoning. As an example, with \textbf{atomic facts} like ``\emph{Alice's son is Bob}'' and ``\emph{Bob's wife is Carol}'', the \textbf{composite query} should be ``\emph{Who is the wife of Alice's son?}''.  Traditional transformer-based LLMs surprisingly struggle to learn this seemingly simple task~\citep{yang2024large,biran2024hopping,balesni2024two, wang2024grokked}, and models often fail to compose two learned facts when the exact composite query was not observed during training. 

The failure was previously attributed to a \textbf{depth-local storage} problem in standard transformers~\citep{biran2024hopping,wang2024grokked}. In this view, a non-recurrent Transformer solves two-hop reasoning through a depth-wise circuit by assigning different roles to layers of different depth. In the example above, earlier layers can recover the first-hop bridge entity, ``\emph{Bob}'', from ``\emph{Alice's son is Bob}'', while later layers must use this bridge to retrieve the second-hop answer ``\emph{Carol}'', from ``\emph{Bob's wife is Carol}''. This depth-wise division of labor requires the second-hop fact to be available in the later layers where the answer is retrieved. However, if that fact is observed during training only as an atomic fact and never in a compositional second-hop role, training provides little direct pressure to make it available at the required depth.

This perspective motivates looped Transformers~\citep{wang2024grokked,zeng2025pretraining,zhu2025scaling,kohli2026loop} as a natural architectural fix. By applying the same Transformer block recurrently, a looped model reuses the same learned parametric memory across reasoning steps: the first loop can recover the bridge, and the next loop can query the same memory to resolve the second hop. However, despite substantially outperforming non-looped Transformers, existing looped-Transformer architectures can still suffer from low training efficiency~\citep{wang2024grokked} and imperfect generalization~\citep{ye2025does} especially in out-of-distribution (OOD) settings, where the constituent atomic facts are observed during training only as standalone facts and never appear in any compositional two-hop training example~\citep{ye2025does}. This raises our motivating questions:
\begin{center}
    \emph{If looping makes the same memory available across hops, why do vanilla looped transformers still fail to compose atomic facts? How do we resolve the remaining issue?}
\end{center}

Based on the two questions, we design a controlled symbolic two-hop reasoning task and conduct mechanistic interpretability analyses on vanilla looped transformers (see Section~\ref{sec:background} and Section~\ref{sec:loop-imperfect}). We identify a new \emph{\textbf{representation}} bottleneck: after the first loop, the bridge entity is often decodable from the continuous hidden vector, yet this vector remains noisy and geometrically misaligned with the clean discrete embedding of the same entity that the next loop would ideally consume. We further show that a training-free realignment intervention nearly closes the performance gap in both ID and the more challenging OOD setting, which reveals a simple design principle: the recurrence should pass not only continuous hidden states, but also discrete embedding-aligned vectors across loops.

Motivated by this principle, we propose \textbf{\name}, a looped Transformer architecture that keeps the memory-reuse benefit of recurrence while adding a decoded embedding channel to reduce loop-wise representation mismatch. \name~achieves near-perfect test accuracy with substantially fewer training steps across symbolic and synthetic-language multi-hop reasoning tasks. Moreover, on real-world pretraining tasks, \name attains lower training loss and stronger performance on different benchmarks compared to looped-transformer baselines, suggesting that our mixed-channel design generalizes to realistic pretraining setups (see \Cref{sec:experiments}).

\begin{figure}
    \centering
    \includegraphics[width=\linewidth]{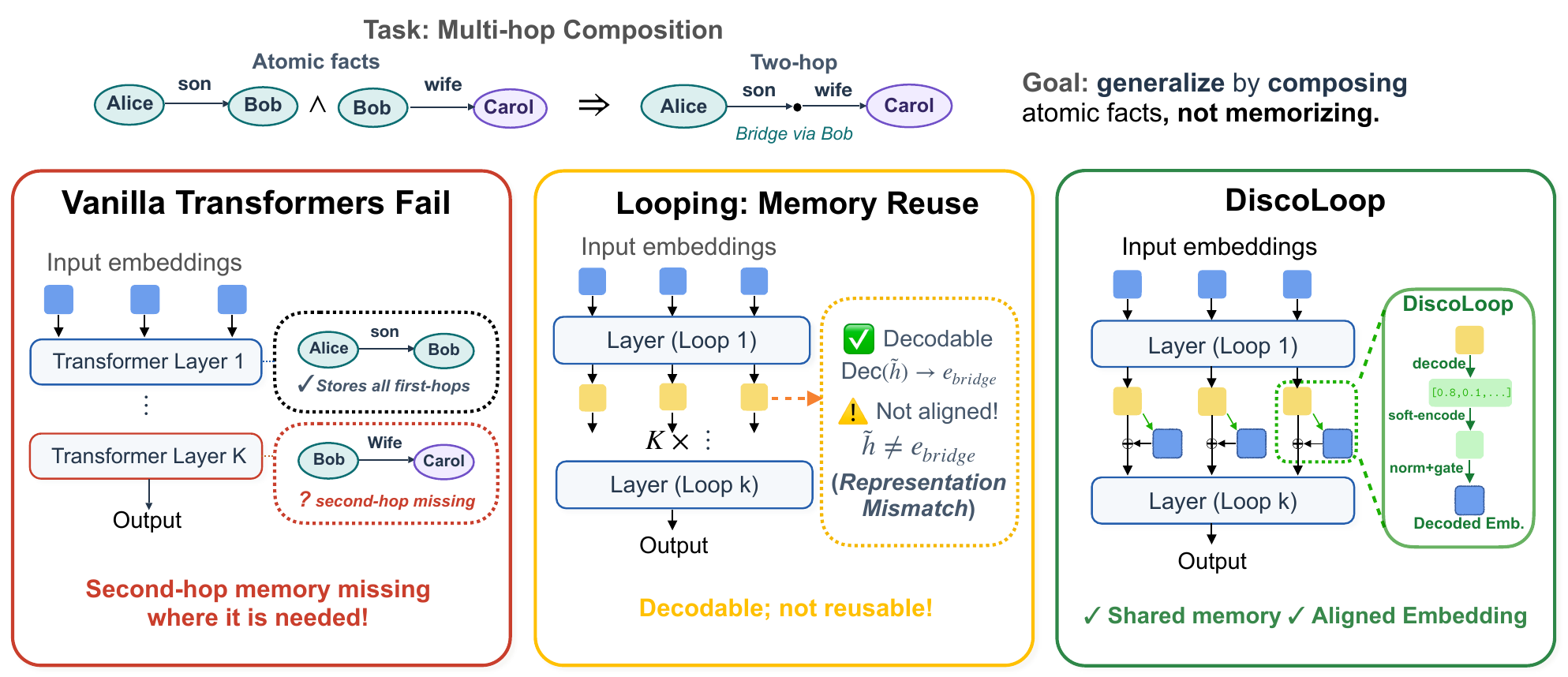}
   \caption{{
\textbf{Overview of our analysis and \name{}.}
The figure summarizes the two-hop reasoning setup, the depth-local storage issue of vanilla non-looped transformers, the representation-mismatch issue that remains in vanilla looped transformers, and our \name architecture that injects a soft decoded embedding into the recurrent residual stream. The rigorous two-hop task formulation is defined in Section~\ref{sec:background}, and the \name recurrence is formalized in Section~\ref{sec:architecture}.
}}
\vspace{-1em}
    \label{fig:overview}
\end{figure}

\subsection{Related work}
\label{subsec:related_work}

\paragraph{Out-of-context and multi-hop reasoning.} Out-of-context reasoning (OCR) refers to models' capability to answer a question where certain key information is missing from the context but learned during training. Previous works show that LLMs can achieve strong OCR performance in certain tasks like factual recall and implication~\citep{feng2024extractive,huang2025generalization}, while they struggle with other OCR tasks, such as the reversal curse~\citep{berglund2023reversal,allen2023physics,zhu2024towards}, in-weight multi-hop reasoning~\citep{yang2024large,biran2024hopping,balesni2024two}, etc. We focus on boosting in-weight multi-hop reasoning capability in this work, which is one of the most important OCR tasks. Prior work~\citep{wang2024grokked} shows that transformers can learn two-hop reasoning through grokking (i.e., an extended period of training far beyond overfitting) when the test data is in distribution (ID), but totally fail for out-of-distribution (OOD) test data (See the rigorous definition of ID and OOD in Section~\ref{sec:background}).  \citet{biran2024hopping,wang2024grokked}  interpreted the failure reason as a depth-local storage issue. \citet{ye2025transformers} showed that the explicit decodability of intermediate entities is insufficient for implicit reasoning, and second-hop generalization requires seeing the relevant atomic fact in the same compositional role during training.

\paragraph{Latent reasoning and looped transformer.} Unlike explicit CoT~\citep{wei2022chain}, latent space reasoning moves the reasoning procedure from the textual token space to a latent space, including discrete latent space~\citep{goyal2023think,wang2023guiding,cheng2024compressed,pfau2024let,su2025token}, continuous latent space~\citep{hao2024training,zhu2025reasoning,zhu2025emergence,gozeten2025continuous,zhang2025soft,butt2025soft}, looped transformer~\citep{saunshi2025reasoning,zeng2025pretraining,zhu2025scaling,kohli2026loop}, etc. In particular, the looping architecture naturally resolves the storage issue of multi-hop reasoning, and the OOD test accuracy increases from 0 to a non-trivial number~\citep{zeng2025pretraining,zhu2025scaling,kohli2026loop} compared to vanilla transformers. However, the OOD test accuracy is still far from perfect. Concurrent to our work, \citet{kohli2026loop} studies implicit in-weight multi-hop reasoning with vanilla looped transformers. Their test split holds out 5\% of atomic facts from training compositions within a single knowledge graph. In contrast, our OOD split uses a separate entity-disjoint graph of comparable size to the training graph and withholds all of its compositions from training, making the two-hop generalization more challenging in both scale and separation.

\section{Task Formulation: Multi-hop Reasoning}\label{sec:background}

\paragraph{Two-hop reasoning.}
We formalize a symbolic abstraction of the two-hop reasoning example in Section~\ref{sec:intro}. Given a knowledge graph $\mathcal{G}=(\mathcal{E},\mathcal{R},\mathcal{F})$ with entity set $\mathcal{E}$, relation set $\mathcal{R}$, and edge set $\mathcal{F}\subseteq \mathcal{E}\times\mathcal{R}\times\mathcal{E}$, an \emph{atomic fact} is a triple $(a, r_1, b)\in\mathcal{F}$, i.e., an edge of $\mathcal{G}$ (e.g., $a=$ Alice, $r_1=$ spouse, $b=$ Bob). A \emph{two-hop question} is a chain $(a, r_1, r_2)$ (e.g., $r_2=$ workplace) whose answer $c$ (e.g., San Francisco) is determined by composing two atomic facts: there exists a unique $b\in\mathcal{E}$ such that $(a,r_1,b)\in\mathcal{F}$ and $(b,r_2,c)\in\mathcal{F}$. We refer to $b$ as the \emph{bridge entity}. Equivalently, two-hop questions are generated by the inference rule
\begin{equation}
\forall a,b,c\in\mathcal{E},\ \forall r_1,r_2\in\mathcal{R},\quad (a, r_1, b)\in\mathcal{F}\ \wedge\ (b, r_2, c)\in\mathcal{F}\ \Longrightarrow\ (a,r_1,r_2,c).
\end{equation}
The construction extends to $k$-hop questions $(a,r_1,\dots,r_k)$ for any $k>1$, but $k=2$ is the most canonical case and thus our main focus throughout the paper. Atomic facts and two-hop questions are presented as separate training examples, and the model is trained to predict the answer $c$ directly from $(a, r_1, r_2)$ without generating the bridge entity $b$. The model is therefore required to store the atomic facts \emph{in weights} and to perform the composition \emph{implicitly} within a single forward pass.

\paragraph{Two-graph design for ID/OOD splits.}
Beyond memorization, we expect that the trained models are able to generalize to compose facts, not only in in-distribution (ID) but also in out-of-distribution (OOD) test sets. We consider two \emph{entity-disjoint} knowledge graphs $\mathcal{G}_A$ and $\mathcal{G}_B$ that \emph{share the same relation vocabulary}. We designate $\mathcal{G}_A$ as the \emph{in-distribution} graph: all atomic facts and a subset of two-hop questions on $\mathcal{G}_A$ are used for training, while the remaining two-hop questions on $\mathcal{G}_A$ form the ID test set $\texttt{test\_id}$. In contrast, $\mathcal{G}_B$ serves as the \emph{out-of-distribution} graph, on which we train \emph{only} its atomic facts; all two-hop questions on $\mathcal{G}_B$ are reserved as the OOD test set $\texttt{test\_ood}$. The crucial asymmetry is that although the model has observed \emph{every} atomic edge of $\mathcal{G}_B$ during training, it has \emph{never} seen a two-hop composition over $\mathcal{G}_B$. Together, the two splits probe whether the composition rule learned on $\mathcal{G}_A$ transfers to held-out questions on the same graph ($\texttt{test\_id}$) and, more stringently, to a disjoint entity set whose atomic facts the model has only seen in isolation ($\texttt{test\_ood}$). This two-graph separation follows the protocol of \citet{wang2024grokked} and makes $\texttt{test\_ood}$ a stricter test of systematic generalization: unlike held-out compositions on $\mathcal{G}_A$, no fact in $\mathcal{G}_B$ has ever appeared as part of a compositional training example, so success requires applying the learned composition rule to facts seen only as standalone facts.

\section{Looping Fixes Storage, but Not Representation}
\label{sec:loop-imperfect}

In this section, we train a vanilla looped transformer on the two-hop reasoning task of Section~\ref{sec:background}. We observe that while looping unlocks implicit two-hop generalization, the test accuracy remains far from perfect, especially out-of-distribution. We then conduct a mechanistic analysis to localize the failure as representational, and find that a simple training-free intervention on the hidden states at the bridge position is enough to nearly close the OOD gap.

\subsection{Training setup}
\label{subsec:loop-setup}

\paragraph{Dataset design.}
We instantiate the two-graph construction of Section~\ref{sec:background} with $|\mathcal{E}_A|=|\mathcal{E}_B|=500$ entities per graph and a shared relation vocabulary of size $|\mathcal{R}|=50$. Every entity has exactly $10$ outgoing edges, each labeled by a distinct relation, so each graph contains $5{,}000$ atomic facts and $50{,}000$ two-hop chains. During training, each entity and each relation is assigned its own dedicated symbolic token (e.g.\ \texttt{<a>}, \texttt{<r\_1>}), so an atomic fact is rendered as a three-token sequence \texttt{<a>\,<r\_1>\,<b>}, and a two-hop question as a four-token sequence \texttt{<a>\,<r\_1>\,<r\_2>\,<c>}. In both cases, the model is trained to predict the final token.  The training set comprises \emph{all} $10{,}000$ atomic facts from both graphs (\texttt{train\_atom}), together with $10{,}000$ two-hop questions (\texttt{train\_id}) drawn uniformly from $\mathcal{G}_A$. We hold out a further $2{,}000$ $\mathcal{G}_A$ chains as the ID test set $\texttt{test\_id}$, and sample $2{,}000$ two-hop questions from $\mathcal{G}_B$ as the OOD test set $\texttt{test\_ood}$. 

\paragraph{Looped Transformer.}
As a starting point, we train a decoder-only Transformer from scratch in a \emph{looping} pipeline. Let $f_\theta$ denote the stack of transformer blocks, and let $\mathbf{W}\in \mathbb{R}^{V\times d}$ be the (tied) embedding matrix, where $V$ is the vocabulary size and $d$ the hidden dimension. Given an input token sequence $x=(x_0,\dots,x_{T-1})$, we form the initial hidden states $\mathbf{H}^{(0)}=\mathbf{W}[x]\in \mathbb{R}^{T\times d}$, and then apply $f_\theta$ recurrently by feeding each iterate's output back in as its own input:
\begin{equation}
\label{eq:loop-recurrence}
\mathbf{H}^{(k+1)} \;=\; f_\theta\!\big(\mathbf{H}^{(k)}\big), \qquad k=0,1,\dots,K-1,
\end{equation}
The next-token logits are read off the final iterate through the same embedding matrix, $\ell = \mathbf{H}^{(K)}\mathbf{W}^{\top}$. We set $K=2$ to match the maximum reasoning depth in the data, giving the model exactly two passes to compose two atomic facts. Intuitively, on a two-hop input $(a,r_1,r_2)$, we expect $f_\theta$ to retrieve the first-hop fact $(a,r_1,b)$ in the first loop and reveal the bridge entity $b$ in the residual stream, and then to use $b$ together with $r_2$ in the second loop to retrieve $(b,r_2,c)$, obtaining  the final answer $c$.

\subsection{Results and mechanistic analysis}
\label{subsec:loop-baseline}

We now train the vanilla looped transformer described in Section~\ref{subsec:loop-setup} on the symbolic dataset and evaluate it across the four splits introduced there. We report accuracy as the fraction of inputs on which the model's argmax-decoded next token at the answer position matches the ground-truth target.

\paragraph{Looped transformers do compose, but imperfectly.}
After $3{,}000$ epochs of training, the looped transformer fits the training set perfectly on both atomic facts and two-hop chains (Table~\ref{tab:loop-baseline}). The key diagnostic is the intermediate \emph{Stage-1} accuracy, obtained by  reading the next-token prediction off the post-first-loop hidden states $\mathbf{H}^{(1)}$ via the LM head $\mathbf{W}$, i.e., applying $f_\theta$ only once ($K=1$) at inference time. Stage-1 atomic accuracy is $100\%$, confirming that atomic facts are perfectly recalled from the weights within a single loop. In contrast, Stage-1 two-hop accuracy on \texttt{train\_id} is only $8.8\%$, far below its perfect Stage-2 (final) score. The model therefore does \emph{not} memorize two-hop answers as shortcuts inside $f_\theta$; instead, the second loop is what performs the composition, exactly as the $K=2$ design intends. Despite this correctly decoupled behavior, the final accuracy is still imperfect: ID test accuracy plateaus at $71.1\%$, and OOD test accuracy at only $8.3\%$. In conclusion, the vanilla looped transformer does the right thing structurally, but its compositional reasoning capability is still far from perfect, particularly outside the training distribution. 

\paragraph{A training-free intervention closes the ID/OOD gap.}
A looped transformer can in principle resolve the depth-local storage issue: the first loop may retrieve the bridge entity before the second-hop fact is needed. However, the second loop must still consume the first-loop representation of this bridge. We therefore test whether making this intermediate representation more embedding-like improves two-hop generalization.

\begin{figure}[h]
\centering
\includegraphics[width=1\linewidth]{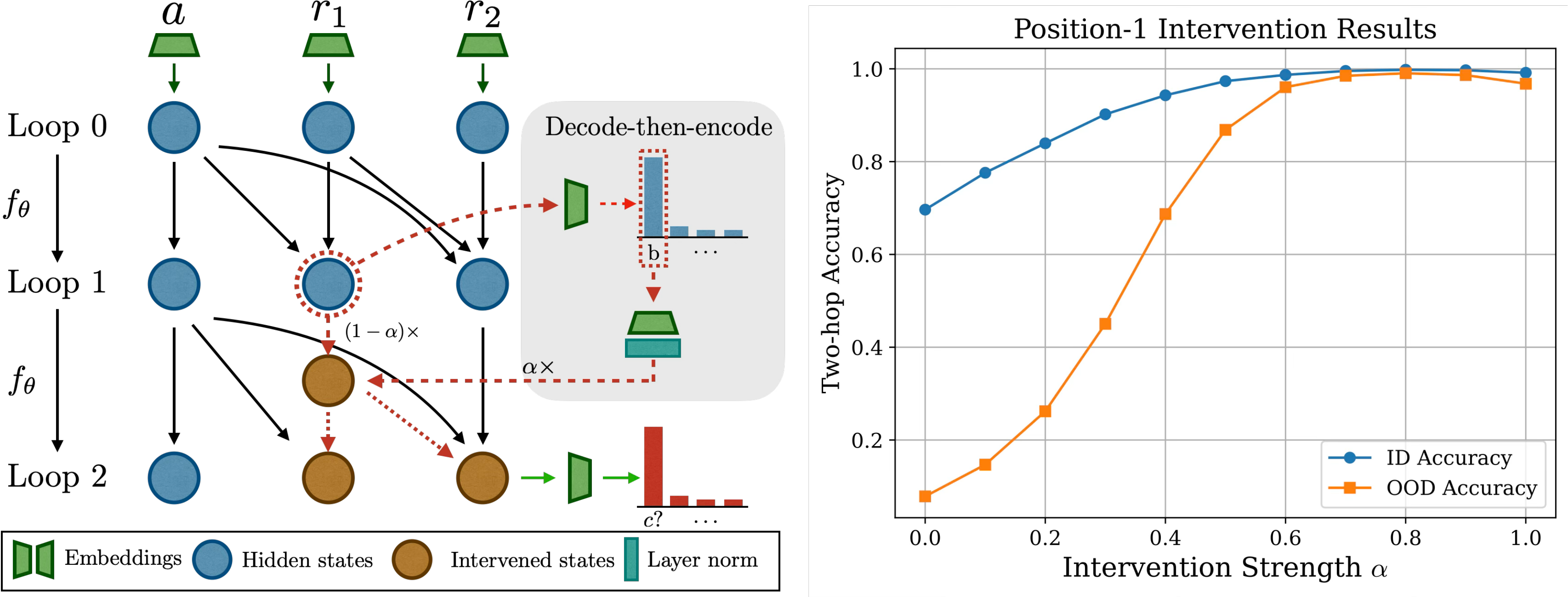}
\caption{\textbf{Left:} the intervention procedure. At $\alpha=0$ the position-$1$ representation is unchanged; at $\alpha=1$ it is replaced entirely by the decoded bridge embedding $\mathrm{Norm}(\mathbf{W}[b_{\rm max}])$. The substitution is applied only at position $1$ and only between the two loops; the rest of the forward pass is untouched. \textbf{Right:} ID and OOD two-hop accuracy as a function of the intervention strength $\alpha$ in Eq.~\eqref{eq:dte}.}
\label{fig:position1-injection}
\end{figure}

We apply a targeted, training-free intervention illustrated in the left panel of Figure~\ref{fig:position1-injection}. Between the first and second loops, at position $1$ only, we replace the hidden state with a convex combination of itself and the normalized embedding of the highest-probability decoded token,
\begin{equation}
\label{eq:dte}
\mathbf{H}^{(1)}_1 \;\leftarrow\; (1-\alpha)\,{\mathbf{H}}^{(1)}_1 \;+\; \alpha\,\mathrm{Norm}(\mathbf{W}[b_{\rm max}]),
\qquad \alpha\in[0,1],
\end{equation}
where $b_{\rm max}=\arg\max(\mathbf{W}\mathbf{H}^{(1)}_1)$ is the top decoded token. Sweeping $\alpha$ produces the right panel: a small intervention ($\alpha=0.1$) already lifts OOD accuracy from $8.3\%$ to $25.9\%$, and at $\alpha\approx 0.5$ both ID and OOD accuracy approach $100\%$, achieving more than a \emph{ten-fold} gain on OOD accuracy.

\paragraph{Decodable bridge, misaligned representation.}
To understand why this intervention helps, we apply a logit lens analysis to the position where the bridge entity $b$ is expected to appear. For every two-hop test input $(a,r_1,r_2)$, we look up the ground-truth bridge $b$ and examine the intermediate hidden state $\mathbf{H}^{(1)}_1$ at position $1$ (the position of $r_1$). We report the intermediate Stage-1 LM-head probability $P(b\mid \mathbf{H}^{(1)}_1)=\mathrm{softmax}(\mathbf{W}\mathbf{H}^{(1)}_1)_{b}$ assigned to the ground-truth bridge token, and the cosine similarity between $\mathbf{H}^{(1)}_1$ and the bridge embedding $\mathbf{W}[b]$.

\begin{table}[h]
\centering
\caption{Results of a vanilla looped transformer on the symbolic dataset. \emph{Stage 1} reads the next-token prediction off the post-first-loop hidden states $\mathbf{H}^{(1)}$; \emph{Stage 2} is the model's actual prediction after both loops.}
\label{tab:loop-results}
\begin{subtable}[t]{0.59\linewidth}
\centering
\caption{Training and test accuracy on the four splits.}
\label{tab:loop-baseline}
\resizebox{\linewidth}{!}{%
\begin{tabular}{lcccc}
\toprule
 & \texttt{train\_atom} & \texttt{train\_id} & \texttt{test\_id} & \texttt{test\_ood} \\
\midrule
Stage 1 & $100.0\%$ & $\phantom{0}8.8\%$ & $\phantom{0}0.9\%$ & $\phantom{0}0.0\%$ \\
Stage 2 & $100.0\%$ & $100.0\%$ & $71.1\%$ & $\phantom{0}8.3\%$ \\
\bottomrule
\end{tabular}%
}
\end{subtable}%
\hfill
\begin{subtable}[t]{0.4\linewidth}
\centering
\caption{Logit lens at position $1$, post-first-loop.}
\label{tab:bridge-analysis}
\resizebox{\linewidth}{!}{%
\begin{tabular}{lcc}
\toprule
 & $P(b\mid \mathbf{H}^{(1)}_1)$ & $\cos(\mathbf{H}^{(1)}_1,\, \mathbf{W}[b])$ \\
\midrule
\texttt{test\_id}  & $1.000$ & $0.327$ \\
\texttt{test\_ood} & $1.000$ & $0.266$ \\
\bottomrule
\end{tabular}%
}
\end{subtable}
\end{table}

As shown in Table~\ref{tab:bridge-analysis}, the model assigns probability essentially $1$ to the correct bridge on both the ID and OOD splits, which indicates that the shared loop block has both pieces needed for composition: the first loop recovers the bridge, and the same block contains the parametric memory to answer $(b, r_2)$ queries in the second loop. This suggests that the remaining failure cannot be simply explained by the depth-local storage issue that limits non-looped Transformers: the bridge is already available before the second loop, and the second-hop fact is ideally retrievable in the second loop since the same block has learned to recall all atomic facts within a single loop.

However, the hidden vector itself is not fully aligned with the bridge embedding direction. The cosine similarity stays around $0.3$, and the OOD cosine is noticeably lower than the ID one. Thus, the second loop receives a noisy continuous mixture rather than a clean copy of the discrete embedding $\mathbf{W}[b]$. Equivalently, the base model $f_\theta$ is asked to consume two qualitatively different input distributions across the two loops: clean token embeddings $\mathbf{W}[x]$ in the first loop, but noisy hidden states $\mathbf{H}^{(1)}$ in the second.

This suggests that the representation mismatch between the noisy continuous hidden state and the clean discrete embedding is a dominant cause of the generalization gap. The larger gap in the OOD case is consistent with this view: OOD entities appear only as endpoints of atomic facts during training and never as intermediate bridges, so $f_\theta$ is never directly pressured to align their post-first-loop hidden states with their embeddings. Therefore, Looping resolves the depth-local storage issue but leaves a representation bottleneck, and mixing in the clean embedding direction at the bridge position is enough to recover near-perfect accuracy. This motivates the architectural change we develop in the next section.

\section{Architecture Design}
\label{sec:architecture}

The intervention in Section~\ref{subsec:loop-baseline} suggests a simple design principle: between loops, the residual stream should carry not just the continuous hidden state computed by $f_\theta$, but also a clean discrete embedding read off through the LM head. The targeted intervention realizes this idea fixed at one position and via a non-differentiable $\arg\max$. We now lift it into a learned, fully differentiable architecture that applies the same principle jointly at every position and every loop.

We propose \textbf{\name}, a looping architecture that augments the residual stream with both \emph{dis}crete embeddings and \emph{co}ntinuous hidden states across the \emph{loop} recurrence. Given a base model $f_{\theta}$, we replace the vanilla recurrence~\eqref{eq:loop-recurrence} by
\begin{equation}
\label{eq:dte-recurrence}
\mathbf{H}^{(k+1)}
\;=\;
f_\theta\!\left(\widetilde{\mathbf{H}}^{(k)}\right),
\qquad
\widetilde{\mathbf{H}}^{(k+1)}
\;=\;
\mathbf{H}^{(k+1)}
+
\alpha^{(k)}
\odot
{\rm RMSNorm}\!\left(\Phi\!\left(\mathbf{H}^{(k+1)}\right)\right),
\end{equation}
for $k=0,\dots,K-1$, with $\widetilde{\mathbf{H}}^{(0)}=\mathbf{W}[x]\in \mathbb{R}^{ T\times d}$. Here $\odot$ denotes row-wise multiplication, so $\alpha^{(k)}=[\alpha^{(k)}_1,\dots,\alpha^{(k)}_{T}]$ is a token-wise gate  along the sequence. $\Phi$ is a soft \emph{decode-then-encode} operator applied independently at each position: \begin{equation} \label{eq:phi-full} \Phi(h) \;=\; \sum_{v=1}^{V}\, p_v(h)\,\mathbf{W}[v], \qquad p_v(h) \;=\; \frac{\exp\big((\mathbf{W} h)_v / \tau\big)}{\sum_{v'=1}^{|V|} \exp\big((\mathbf{W} h)_{v'}/\tau\big)}, \end{equation} with temperature $\tau>0$. Intuitively, $\Phi(h)$ is a weighted sum of token embeddings whose weights are the tempered next-token probabilities obtained by reading the LM head off $h$, which is the differentiable analogue of the hard substitution $\mathbf{W}[v_{\rm max}]$ used in Section~\ref{subsec:loop-baseline}. Adding $\alpha^{(k)}\odot\mathrm{RMSNorm}(\Phi(\mathbf{H}^{(k+1)}))$ to the residual stream therefore injects a \emph{discrete embedding channel} alongside the continuous hidden states, with a gate $\alpha^{(k)}$ controlling the injection strength.

\paragraph{Gate parametrization.}
 We set $\alpha^{(K-1)}\equiv \mathbf{0}_T$, so the final prediction is still read directly from $\mathbf{H}^{(K)}$ by the LM head. For $k=0,\dots,K-2$, we consider two parametrizations of $\alpha^{(k)}$. We can either set a \emph{fixed} $\alpha^{(k)} \equiv \alpha^{\star}\mathbf{1}_{T}$ with $\alpha^{\star}>0$ for all $k$, or design a \emph{learnable} $\alpha^{(k)}$ by 
introducing a token-wise gate through a sigmoid activation $\sigma$:
\begin{equation}
\label{eq:gate-dynamic}
\alpha^{(k)}_t \;=\; \sigma\Big(\big\langle\mathbf{w}_{\alpha},\, \Phi\big(\mathbf{H}^{(k+1)}_t\big)\big\rangle + b_{\alpha}\Big),
\qquad \mathbf{w}_{\alpha}\in\mathbb{R}^d,\ b_{\alpha}\in\mathbb{R}.
\end{equation}
The gate is conditioned on the decoded content $\Phi(\mathbf{H}^{(k+1)}_t)$ at each token, so the model can inject the discrete embedding more strongly at positions where the decoded distribution is informative and turn the injection off elsewhere. The learnable variant introduces only $d+1$ extra parameters in total (a single vector $\mathbf{w}_\alpha$ and bias $b_\alpha$ shared across loops and positions), which is negligible relative to the size of $f_\theta$.

\section{Experiments}
\label{sec:experiments}

We evaluate \name{} in three settings of increasing realism. Section~\ref{subsec:experiments-symbolic} first revisits the symbolic two-graph task from Section~\ref{subsec:loop-setup}. Section~\ref{subsec:experiments-nl} then studies a synthetic natural-language version of the same task in two question formats, testing whether the same effect persists beyond pure symbolic tokenization. Section~\ref{subsec:experiments-lm} evaluates practical language modeling, asking whether the \name also improves standard pretraining beyond the controlled multi-hop datasets.

\subsection{Symbolic two-graph dataset}
\label{subsec:experiments-symbolic}

We re-use the symbolic dataset and training setup of Section~\ref{subsec:loop-setup}. We compare \name{} with two baselines: (i) a \emph{vanilla looped transformer} of Eq.~\eqref{eq:loop-recurrence}, which shares the same base model $f_\theta$ as \name{} and differs only in whether the discrete embedding channel is added, and (ii) a \emph{non-looped transformer} obtained by stacking $K$ copies of $f_\theta$ as untied layers. This non-looped baseline matches the FLOPs of the looped models while strictly upper-bounding their parameter count. For \name{}, we use a fixed gate $\alpha^{\star}=1$ and temperature $\tau=1$. All other training hyperparameters are kept identical across the three models.

\begin{figure}[h]
    \centering
    \includegraphics[width=\linewidth]{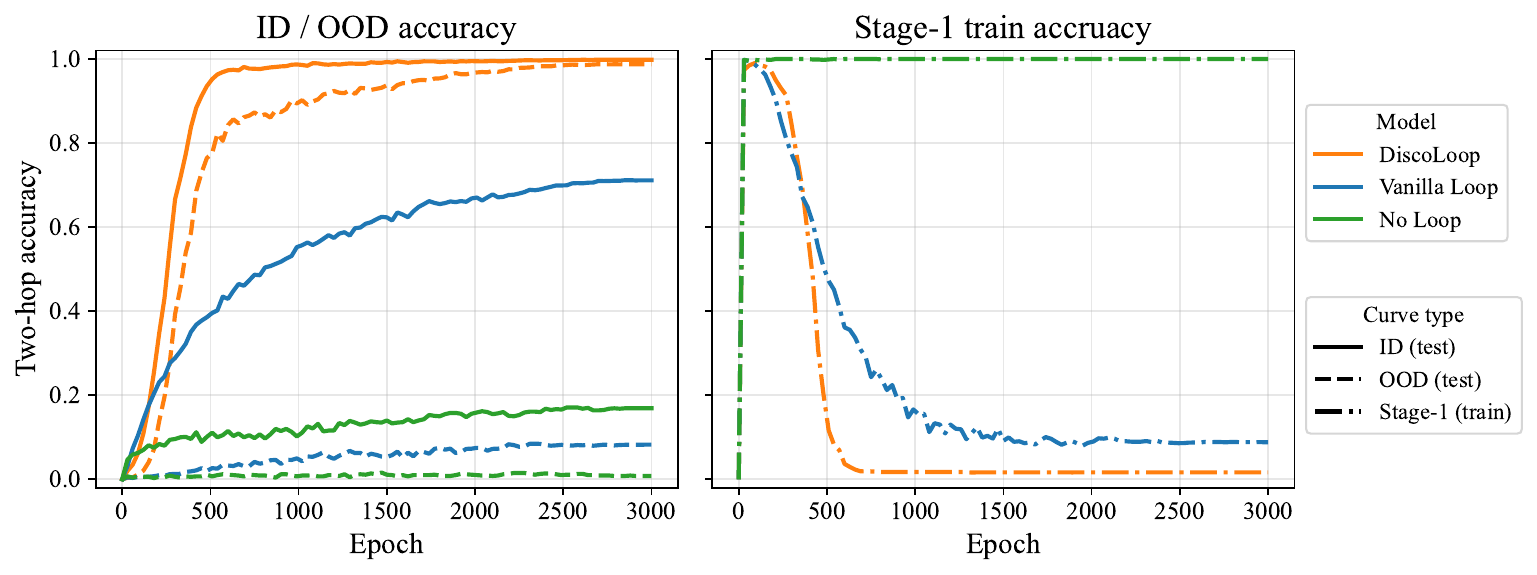}
    \caption{Two-hop accuracy along training on the symbolic dataset for the non-looped transformer, the vanilla looped transformer, and \name{}. \textbf{Left:} Final two-hop accuracy on the held-out ID test set (solid) and the OOD test set (dashed). \textbf{Right:} Stage-1 two-hop training accuracy, obtained by reading the LM head after the first loop. Since the non-looped transformer has no recurrent stages, we report its final two-hop training accuracy as a reference.}
    \label{fig:exp-curves-symbolic}
\end{figure}

\paragraph{\name{} dominates both baselines.}
According to the left panel of Figure~\ref{fig:exp-curves-symbolic}, \name{} reaches near-perfect accuracy on both the ID and OOD test sets by the end of training, while the vanilla looped transformer plateaus at roughly $70\%$ ID and below $10\%$ OOD. The non-looped baseline performs the worst despite using $K\!\times$  parameters: only below $20\%$ ID accuracy essentially $0\%$ OOD. \name{} also demonstrates substantially better training efficiency: we observe a \textbf{sharp phase transition} around epoch $500$, after which both ID and OOD test accuracy already exceed $80\%$.

\paragraph{Stronger inductive bias towards compositional reasoning.}
The right panel of Figure~\ref{fig:exp-curves-symbolic} plots the Stage-1 two-hop \emph{training} accuracy, obtained by reading the LM head off the post-first-loop hidden states $\mathbf{H}^{(1)}$ as described in Section~\ref{subsec:loop-baseline}. Early in training, both looped models behave like the non-looped baseline: Stage-1 training accuracy quickly hits $100\%$ while ID and OOD test accuracy stay near zero, indicating that two-hop answers are being memorized directly in the weights of $f_\theta$. During the phase transition, the Stage-1 training accuracy \emph{collapses toward zero} just as the test accuracies \emph{jump up}. Once the model can no longer answer two-hop training questions from a single loop, it is forced to offload composition to the second loop. This is the looping architecture's inductive bias against single-loop memorization shortcuts, leading to a more generalizable solution.

Crucially, the Stage-1 collapse is \emph{earlier and sharper} for \name{} than for the vanilla looped baseline, and its test accuracy rises earlier and converges higher. By injecting a cleaner signal via the discrete embedding channel $\Phi$, \name{} strengthens this inductive bias, pushing the model toward looping-based composition more aggressively, thus accelerating the phase transition and improving final generalization.

\paragraph{Beyond two-hop.}
The benefits of \name{} are not limited to two-hop reasoning. We additionally evaluate it on a \emph{three-hop} extension of the symbolic dataset and find that it again converges to near-perfect ID accuracy and substantial OOD generalization, while the vanilla looped transformer and the non-looped baseline both fail completely on the three-hop OOD split. Due to space constraints, we defer the full setup and results to Appendix~\ref{appendix:3hop}.

\subsection{Synthetic natural-language dataset}
\label{subsec:experiments-nl}

To check that the architectural finding is not an artifact of symbolic tokenization, we re-run the same task on a synthetic natural-language version where atomic facts and two-hop questions are verbalized in English. We again generate a two-graph dataset as described in Section~\ref{sec:background}. Instead of a pure symbolic sequence, each atomic fact is rendered as ``\texttt{Finley's teacher is Anya}'', and we consider two natural verbalizations of the two-hop query that differ in the order of the relation chain:
\begin{itemize}
\item \emph{Direct}: ``\texttt{Finley's teacher's wife is \textbf{Charlie}}''. 
\item \emph{Reverse}: ``\texttt{Who is the wife of the teacher of Finley?\textbackslash nAnswer: \textbf{Charlie}}''. 
\end{itemize}
The two formats share the same atomic facts and the same answer for every two-hop chain; only the surface form of the question differs. The reverse format is closer to natural-language QA but requires the model to compose relations in the opposite order to that in which they appear in the input.

We again use $K=2$ loops for both \name{} and the vanilla looped transformer, and the corresponding non-looped baseline stacks two copies of $f_\theta$ as untied layers. We train all models for $5000$ epochs, with the remaining training setup identical to the symbolic experiments. We use a learnable gate $\alpha$ for \name{} on both synthetic-language datasets.

\begin{figure}[h]
\centering
\includegraphics[width=1\linewidth]{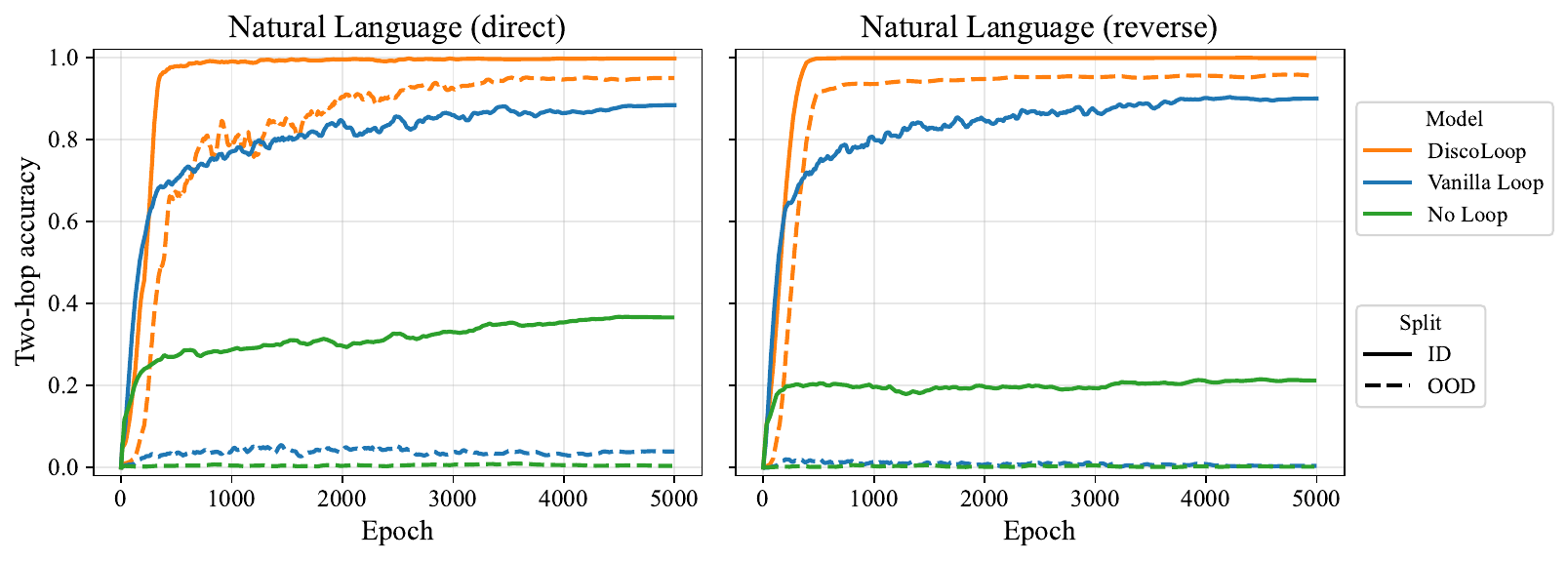}
\caption{Two-hop accuracy on the two synthetic-language datasets (\emph{direct} and \emph{reverse} formats), comparing the non-looped transformer, the vanilla looped transformer, and \name{} on the ID test set (solid) and the OOD test set (dashed). All models are trained under the same optimization recipe used in the symbolic setting, except that the total number of training epochs is $5000$ rather than $3000$.}
\label{fig:exp-curves-nl}
\end{figure}

Figure~\ref{fig:exp-curves-nl} shows that the language-setting curves mirror the symbolic case. For both formats, \name{} outperforms both baselines in accuracy and training efficiency. \name{} reaches near $100\%$ ID accuracy and around $95\%$ OOD accuracy, while the vanilla looped transformer reaches around $90\%$ ID accuracy but fails OOD, especially in the reverse format. The non-looped baseline reaches only around $40\%$ ID accuracy on the direct format and $20\%$ on the reverse format, and essentially $0\%$ OOD accuracy on both. \name{} also generalizes perfectly on the ID test set well before the end of training, also around epoch $500$, making it substantially more training-efficient than the baselines. We observe the same inductive-bias phenomenon as in the symbolic case, which we visualize in Appendix~\ref{app:nl-stage1}.
\subsection{Language modeling}
\label{subsec:experiments-lm}

We next test whether the benefits of \name{} extend beyond controlled synthetic tasks to practical language modeling. Besides the vanilla loop baseline, we compare with PonderLM~\citep{zeng2025pretraining}, a related recurrent architecture that also maps intermediate predictions back into embedding space. The key difference is how the recurrent signal is carried: PonderLM maintains the recurrence only in embedding space, while \name{} preserves the continuous hidden state and injects the decoded embedding as an additional discrete channel. Thus this comparison tests whether keeping both channels improves over both hidden-state-only looping and embedding-only looping.

We pretrain three $440$M models on $20$B tokens using a $6{:}4$ mixture of FineWeb-Edu and FineMath~\citep{niklaus2026smoldata,allal2025smollm2smolgoesbig}. The FineMath component makes the corpus more math- and reasoning-heavy than pure web text. All models share the same model backbone as \citet{zhu2025scaling} with hidden dimension $d=1024$, $24$ layers, loop step $4$, and tied input-output embeddings. We keep the training recipe fixed and vary only the recurrent computation mechanism. For a controlled comparison, all recurrent models use four total applications of the backbone. Additional training details are provided in Appendix~\ref{appendix:lm-pretraining-details}.
\begin{table*}[t]
\centering
\small
\setlength{\tabcolsep}{3.8pt}
\renewcommand{\arraystretch}{1.18}
\caption{
Zero-shot evaluation scores for the $440$M pretraining runs using the backbone architecture of \citet{zhu2025scaling}, with 24 layers, loop step $4$, and tied input-output embeddings. For multiple-choice tasks, we report length-normalized accuracy to reduce length bias. Thus ARC-C, ARC-E, HellaSwag, PIQA, and SciQ use normalized accuracy; LAMBADA and RACE use accuracy. Bold indicates the best score for each benchmark. \name is best or tied on six of the seven benchmarks.
}
\label{tab:loop-cot-final2-results}
\begin{threeparttable}
\begin{tabular}{lcccccccc}
\toprule
Setting & ARC-C & ARC-E & HellaSwag & LAMBADA & PIQA & RACE & SciQ & Avg. \\
\midrule
Vanilla loop 
& 30.5
& 57.2
& \textbf{44.2}
& 35.1
& 67.4
& 31.0
& 79.5
& 49.3 \\
PonderLM 
& 31.7
& 55.6
& 43.9
& \textbf{38.0}
& 67.7
& 31.2
& 80.5
& 49.8 \\
\midrule
\textbf{\name} (Ours) 
& \textbf{32.6}
& \textbf{57.6}
& \textbf{44.2}
& 37.6
& \textbf{68.4}
& \textbf{31.3}
& \textbf{81.5}
& \textbf{50.5} \\
\bottomrule
\end{tabular}
\end{threeparttable}
\end{table*}

\paragraph{\name improves the language modeling.}
Table~\ref{tab:loop-cot-final2-results} reports zero-shot evaluation on seven standard language-modeling benchmarks. \name obtains the best average score, improving over the vanilla Loop baseline from $49.3$ to $50.5$, and also outperforming PonderLM's average score of $49.8$. \name is best or tied on six of the seven benchmarks, with the largest numerical improvements over Loop on ARC-C, LAMBADA, and SciQ. PonderLM remains strongest on LAMBADA. When aggregating across benchmarks, \name achieves the highest average score, suggesting that the looped reasoning mechanism yields broad gains rather than improvements confined to a small subset of tasks. This indicates that the mechanism does not merely solve the synthetic compositional task, but also transfers to standard language model pretraining. We also present the training loss curves in Appendix~\ref{appendix:lm-pretraining-details}. \name surpasses the vanilla loop after training on roughly 13B tokens and maintains this advantage through the end of training.

\section{Conclusion and Discussion}
\label{sec:conclusion}

We have studied implicit in-weight multi-hop reasoning through the lens of looped transformers. Our contribution has been two-fold. First, we have provided a mechanistic explanation for why vanilla looped transformers do not fully solve the task. Although the bridge entity is often already decodable after the first loop, the hidden state carrying it is not well aligned with the clean token embedding that the next loop would naturally consume. A simple training-free intervention that injects the decoded bridge embedding has nearly closed the ID/OOD gap, showing that representation mismatch is a central bottleneck. Second, motivated by this finding, we have proposed \name{}, which augments the loop recurrence with both a continuous hidden-state channel and a decoded discrete-embedding channel. \name{} has achieved strong performance on symbolic and synthetic-language two-hop reasoning tasks, improving both training efficiency and OOD generalization, and has also shown gains in our open-domain pretraining experiments.

Looking forward, it would be interesting to test whether \name{} length-generalizes: trained on questions with at most $k$ hops, can it extrapolate to questions with more than $k$ hops? This would be a stronger test of whether the architecture learns a more generalizable compositional procedure rather than a depth-specific strategy. Moreover, our pretraining experiments have been limited to moderate-scale models because of compute constraints, so larger-scale verification remains important. Beyond training looped models from scratch, an important next step is to study whether the same looping recipe can be introduced through continual pretraining of larger pretrained models. This would test whether existing pretrained non-looped models can acquire the discrete-continuous recurrence without full retraining, and whether the benefits of \name{} persist at more realistic model scales.

\section{Acknowledgements}
HF and JDL acknowledge support of  NSF IIS 2107304,  NSF CCF 2212262, NSF CAREER Award 2540142, NSF 2546544, NSF CCF 2019844 and ONR N00014-24-1-2639. SM acknowledges support of ONR N00014-24-1-2639.

\bibliography{references}
\bibliographystyle{plainnat}

\newpage 

\appendix

\section{Additional experiments and training details on synthetic datasets}
\subsection{Training details}\label{sec:implementation and compute}
We implement the vanilla non-looped transformers, looped-transformers and \name all through PyTorch and Huggingface Transformers \citep{paszke2019pytorch,wolf2019huggingface}. All the synthetic experiments are done on 8 NVIDIA H200 GPUs within 168 hours at maximum.
\subsection{Three-hop reasoning experiment}
\label{appendix:3hop}

To check that the architectural advantage of \name is not specific to two-hop composition, we run the same comparison on a three-hop extension of the symbolic two-graph task.

\paragraph{Three-hop questions.}
Following the symbolic abstraction in Section~\ref{sec:background}, a \emph{three-hop question} is a chain $(a, r_1, r_2, r_3)$ whose answer $d$ is determined by composing three atomic facts: there exist unique $b, c \in \mathcal{E}$ such that $(a, r_1, b),\,(b, r_2, c),\,(c, r_3, d) \in \mathcal{F}$. We refer to $b$ and $c$ as the two bridge entities, and the three-hop input is rendered as a five-token sequence \texttt{<a>\,<r\_1>\,<r\_2>\,<r\_3>\,<d>}, with the model trained to predict the final answer token.

\paragraph{Dataset.}
We re-use the two-graph construction of Section~\ref{subsec:loop-setup}: $|\mathcal{E}_A| = |\mathcal{E}_B| = 500$ entities per graph, a shared relation vocabulary of size $|\mathcal{R}| = 50$, and out-degree $10$, so each graph contains $5{,}000$ atomic facts. The training set is composed of \emph{all} $10{,}000$ atomic facts from both graphs, $5{,}000$ two-hop questions, and $5{,}000$ three-hop questions, both sampled from $\mathcal{G}_A$. We evaluate on four held-out test splits of $1{,}000$ chains each: two-hop ID and three-hop ID, sampled from $\mathcal{G}_A$ disjoint from training; and two-hop OOD and three-hop OOD, sampled from $\mathcal{G}_B$. Since no compositional question over $\mathcal{G}_B$ is ever seen in training, both OOD splits cleanly probe systematic compositional generalization at the corresponding reasoning depth.

\paragraph{Models and training.}
We set $K=3$ loops for both the vanilla looped transformer and \name, matching the maximum reasoning depth in the data. The non-looped baseline correspondingly stacks three copies of $f_\theta$ as untied layers. We use a fixed gating $\alpha=1$ for \name. All other hyperparameters (base model, optimizer, learning rate, batch size, schedule) follow Section~\ref{subsec:loop-setup}.

\begin{figure}[h]
    \centering
    \includegraphics[width=\linewidth]{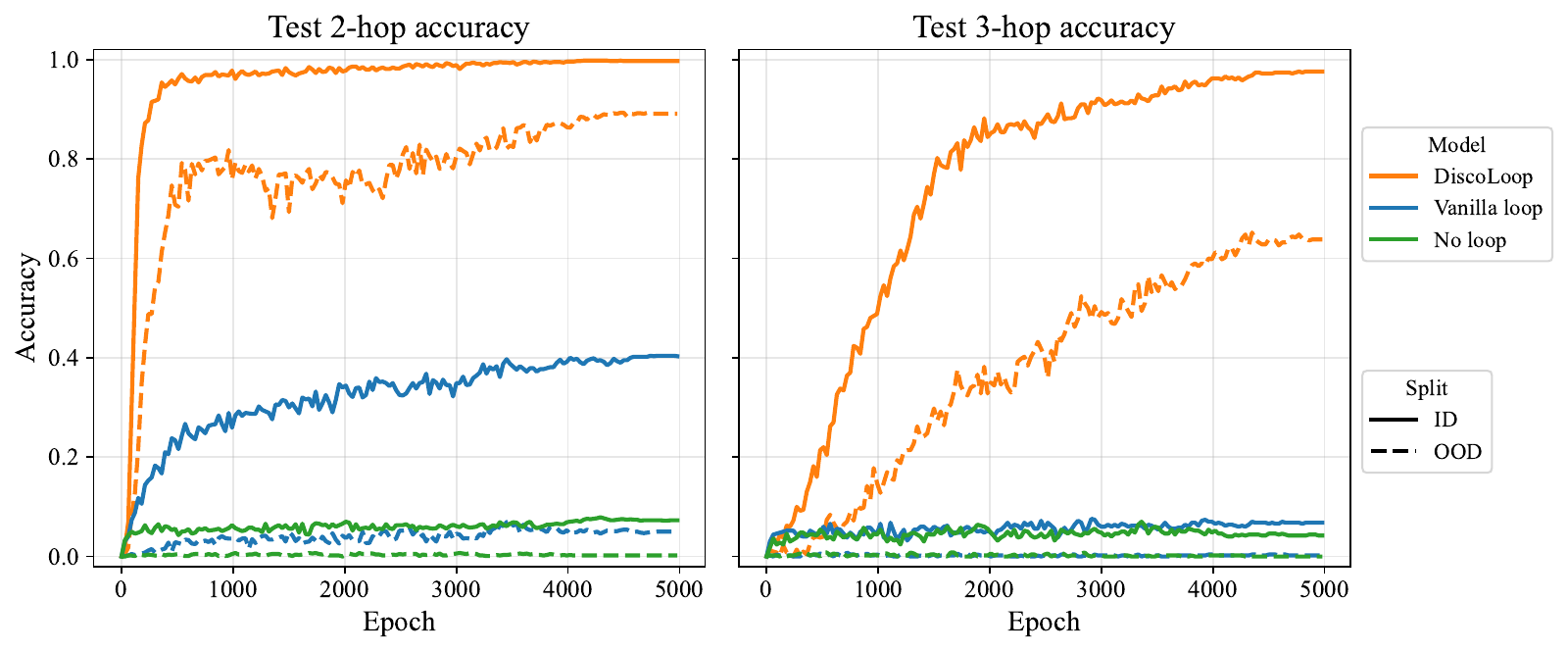}
    \caption{Test accuracy on the three-hop symbolic dataset for the non-looped transformer, the vanilla looped transformer, and \name. Solid lines: ID test set. Dashed lines: OOD test set. \textbf{Left:} two-hop test accuracy. \textbf{Right:} three-hop test accuracy.}
    \label{fig:3hop-results}
\end{figure}

\paragraph{Results.}
Figure~\ref{fig:3hop-results} shows test accuracy on the four splits over training. The curves resemble the two-hop case but with a more pronounced separation. On the two-hop splits (left panel), \name reaches near-perfect ID accuracy and around $90\%$ OOD, while the vanilla looped transformer plateaus near $40\%$ ID and $5\%$ OOD, and the non-looped baseline fails on both. On the three-hop splits (right panel), \name again converges to near-perfect ID accuracy and reaches roughly $65\%$ OOD. Strikingly, both the vanilla looped transformer and the non-looped baseline remain near zero on the three-hop OOD split throughout training.

In conclusion,  \name's advantage is not specific to two-hop composition: the same mechanism continues to accelerate training and improve generalization at deeper reasoning depths. Moreover, the gap between \name and the baselines \emph{widens} with depth. The vanilla looped transformer can still partially solve two-hop ID, but almost fails on three-hop, where each additional hop requires another clean handoff between loops. This is consistent with the mechanistic story in Section~\ref{subsec:loop-baseline}: representational misalignment at the bridge position compounds across hops in the vanilla model, while \name's discrete channel suppresses it at every hop, unlocking compositional generalization that is otherwise unreachable.

\subsection{Stage-1 inductive bias on the natural-language datasets}
\label{app:nl-stage1}

\begin{figure}[h]
\centering
\includegraphics[width=0.95\linewidth]{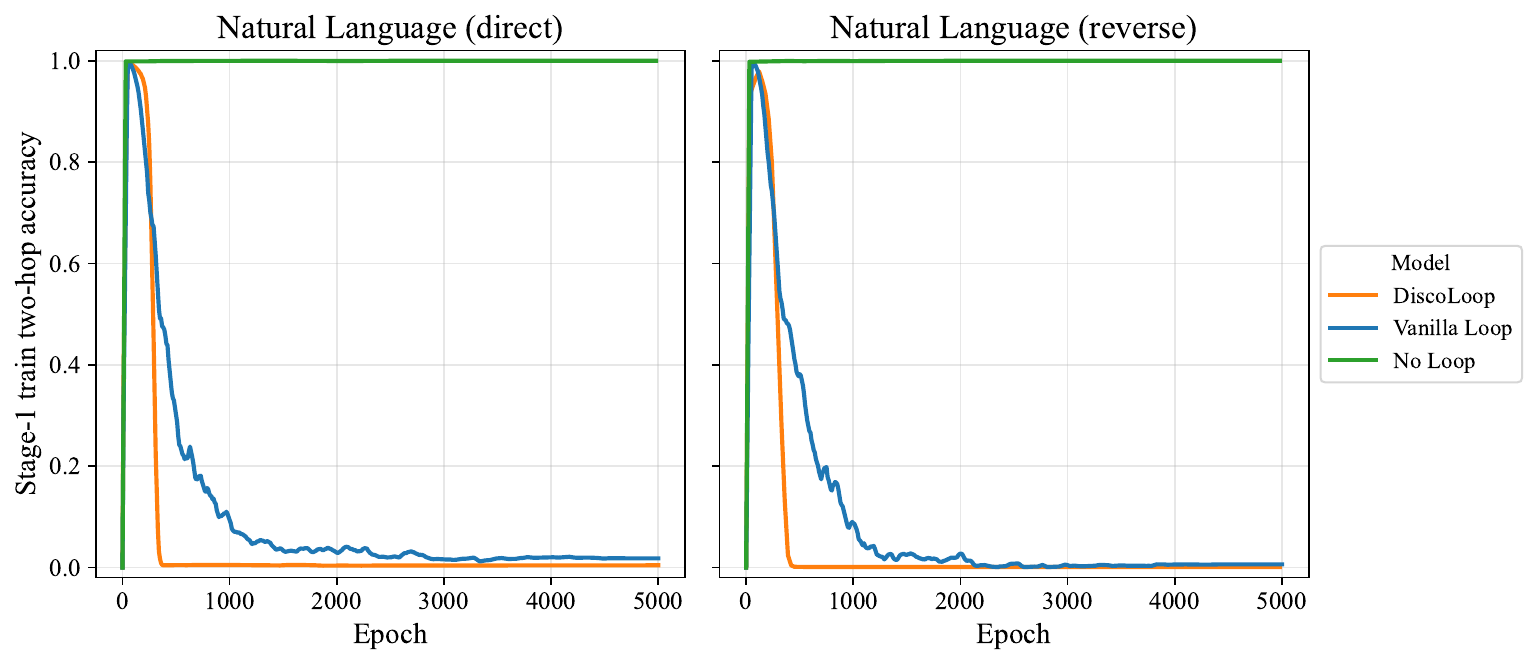}
\caption{Stage-1 two-hop training accuracy on the two synthetic natural-language datasets (\emph{direct} and \emph{reverse} formats), for the non-looped transformer, the vanilla looped transformer, and \name. As in the symbolic case, both looped models suppress their Stage-1 training accuracy as composition is offloaded from $f_\theta$ to the looping recurrence; the suppression is markedly sharper for \name than for the vanilla looped transformer. The non-looped baseline retains $\sim\!100\%$ throughout, having no second loop to offload composition to.}
\label{fig:nl-stage1}
\end{figure}

Figure~\ref{fig:nl-stage1} plots the Stage-1 two-hop training accuracy on the two natural-language datasets, computed exactly as in the symbolic case (Section~\ref{subsec:experiments-symbolic}). The pattern mirrors that case: both looped models initially fit the two-hop training set within a single loop and then sharply suppress this single-loop accuracy once composition is offloaded to the looping recurrence. The collapse is remarkably sharper for \name than for the vanilla looped transformer---\name drops to near-zero Stage-1 accuracy within roughly $400$ epochs, while the vanilla baseline takes more than $2{,}000$ epochs to do the same. The non-looped baseline retains $\sim\!100\%$ Stage-1 accuracy throughout, since it has no second loop to offload composition to. The natural-language results therefore reproduce the same inductive-bias phenomenon observed in the symbolic setting, with \name again strengthening it.

\section{Language modeling pretraining details}
\label{appendix:lm-pretraining-details}

This section gives the full training and evaluation details for the open-domain language-modeling experiments in Section~\ref{subsec:experiments-lm}. We adopt the Ouro architecture from \citet{zhu2025scaling} because it provides a strong and established design for loop transformers. The recorded launcher configuration corresponds to the vanilla loop baseline. \name{} and PonderLM use the same architecture scale, data mixture, tokenizer, optimizer, sequence length, and distributed training setup, changing only the recurrent computation mechanism. All recurrent models apply the backbone four times, i.e., $K=4$. 

For \name{}, we use a top-$k$ approximation to the decode-then-encode operator $\Phi$ in Eq.~\eqref{eq:phi-full}. For a hidden state $h$, let
\[
\mathcal{T}_k(h)
    =
    \operatorname{TopK}\big(\{p_v(h)\}_{v=1}^{V}\big)
    \subseteq [V]
\]
denote the set of $k$ tokens with the largest probabilities under the LM-head distribution $p(h)$. We then replace $\Phi$ by
\[
\Phi_k(h)
    =
    \sum_{v\in \mathcal{T}_k(h)}
    p_v(h)
    \mathbf{W}[v],
\]
 where we mask all logits outside the top-$k$ set and then computing the embedding average on this truncated support. This reduces compute and memory overhead, since most probabilities $p_v(h)$ are negligible in practice. We find that $k=128$ is sufficient to retain essentially all nontrivial probability mass in our pretraining runs.

For PonderLM, four backbone applications correspond to three ponder steps under the convention of the original PonderLM formulation, where the final backbone application is treated as the standard final forward pass rather than a ponder step.

\paragraph{Architecture.}
We use a looped transformer backbone \cite{zhu2025scaling}~with hidden size $1024$, $24$ layers, and $16$ attention heads. The recurrent computation uses four total applications of the backbone. The context length is $8192$, with maximum position embeddings set to $16384$ and RoPE base $\theta=10^6$. The model uses SwiGLU activations, tied input/output embeddings, no QKV bias, and vocabulary size $129{,}280$.

\paragraph{Training data.}
Unless otherwise specified, all runs use the same two-source pretraining mixture: $60\%$ FineWeb-Edu sample data and $40\%$ FineMath-4plus. Both sources are loaded as streaming parquet datasets. We use the DeepSeek-V3-0324 tokenizer.

\paragraph{Optimization.}
Models are trained with AdamW and Muon enabled. We use learning rate $3\times 10^{-3}$, AdamW $\epsilon=10^{-15}$, Muon momentum $0.95$, gradient clipping at norm $1.0$, and a linear learning-rate schedule with $1000$ warmup steps and final absolute learning rate $10^{-5}$. The effective batch size is $128$, the sequence length is $8192$.

\paragraph{Token budget and distributed setup.}
The Loop baseline is trained with a target budget of $20$B tokens. With $2$ nodes, $8$ GPUs per node, batch size $8$, sequence length $8192$, and gradient accumulation $1$, each step processes $ 1{,}048{,}576$ tokens. This gives $19{,}074$ optimization steps, corresponding to $20{,}000{,}538{,}624$ total training tokens.

\paragraph{Evaluation.}
We evaluate checkpoints using \texttt{lm-evaluation-harness} in the zero-shot setting. The reported benchmarks are ARC-C, ARC-E, HellaSwag, LAMBADA, PIQA, RACE, and SciQ. For multiple-choice tasks where \texttt{lm-evaluation-harness} \citep{eval-harness} provides \texttt{acc\_norm}, we report \texttt{acc\_norm}; otherwise we report \texttt{acc}. 

\begin{figure}[t]
    \centering
    \includegraphics[width=0.8\linewidth]{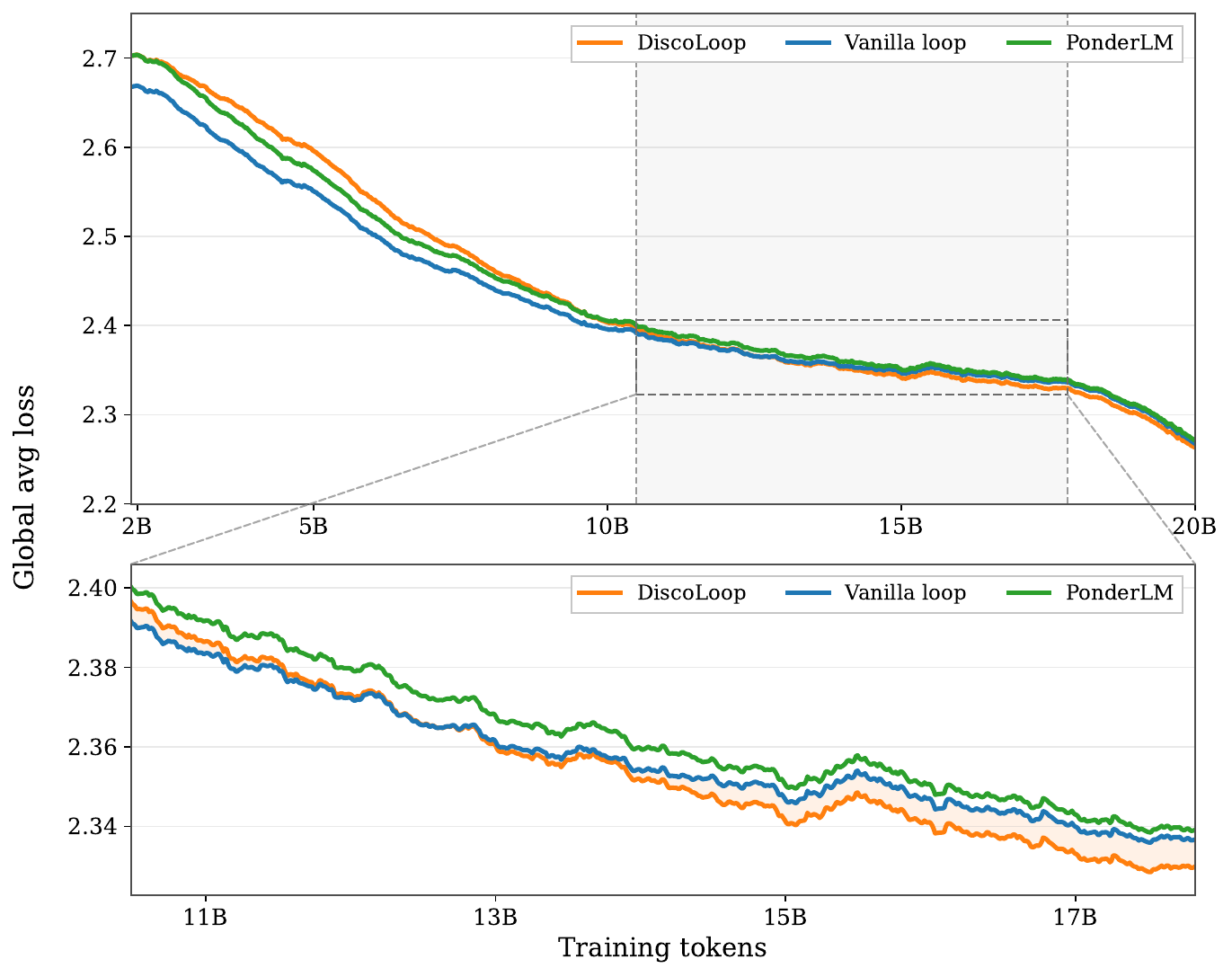}
    \caption{\textbf{Pretraining loss curves for the language modeling.}
    Global average loss versus training tokens for \name, Vanilla loop, and PonderLM over the full $20$B-token training budget ($19{,}074$ optimization steps). For readability, curves are smoothed with an exponential moving average. The upper panel shows the full training trajectory, while the lower panel zooms into the late-training regime from roughly $10$B to $17$B tokens; the dashed box and connector lines in the upper panel indicate the zoomed interval. Vanilla loop leads early in training, but \name overtakes it in the second half of training and attains the lowest loss in the zoomed region, with a peak separation from Vanilla loop on the order of $10^{-2}$. PonderLM also starts above Vanilla loop and, although it steadily improves, remains above \name throughout the zoomed window.}
    \label{fig:d1024-loss}
\end{figure}

\paragraph{Training loss curves.}
Figure~\ref{fig:d1024-loss} shows the pretraining global average loss for the three settings: \name, Vanilla loop, and PonderLM. Early in training, Vanilla loop has the lowest loss, while \name and PonderLM start slightly higher. In the second half of training, however, \name undergoes a clear transition and overtakes Vanilla loop, becoming the best-performing run in the late stage. In the zoomed region, this advantage is sustained and reaches roughly $10^{-2}$ in global average loss at its largest, while PonderLM remains above \name throughout this phase.

\end{document}